\documentclass[journal,comsoc]{IEEEtran}

\usepackage[T1]{fontenc}% optional T1 font encoding
\usepackage{graphicx}

\ifCLASSINFOpdf
\else

\fi
\usepackage{amsmath}

\usepackage[cmintegrals]{newtxmath}

\begin{document}

\title{Offline signature authenticity verification through \\ unambiguously connected skeleton segments}

\author{Jugurta~Montalv\~ao,
        Luiz~Miranda,
        and~J\^anio~Canuto}% <-this % stops a space

% The paper headers
%\markboth{Journal of Communication and Information Systems}%
%{Shell \MakeLowercase{\textit{et al.}}: Offline signature authenticity verification through unambiguously connected skeleton segments}

% make the title area
\maketitle

% As a general rule, do not put math, special symbols or citations
% in the abstract or keywords.
\begin{abstract}
A method for offline signature verification is presented in this paper. It is based on the segmentation of the signature skeleton (through standard image skeletonization) into unambiguous sequences of points, or unambiguously connected skeleton segments corresponding to vectorial representations of signature portions. These segments are assumed to be the fundamental carriers of useful information for authenticity verification, and are compactly encoded as sets of 9 scalars (4 sampled coordinates and 1 length measure). Thus signature authenticity is inferred through Euclidean distance based comparisons between pairs of such compact representations. The average performance of this method is evaluated through experiments with offline versions of signatures from the MCYT-100 database. For comparison purposes, three other approaches are applied to the same set of signatures, namely: (1) a straightforward approach based on Dynamic Time Warping distances between segments, (2) a published method by \cite{shanker2007}, also based on DTW, and (3) the average human performance under equivalent experimental protocol. Results suggest that if human performance is taken as a goal for automatic verification, then we should discard signature shape details to approach this goal. Moreover, our best result -- close to human performance -- was obtained by the simplest strategy, where equal weights were given to segment shape and length.\end{abstract}

% Note that keywords are not normally used for peerreview papers.

\IEEEpeerreviewmaketitle

\section{Introduction}\label{Intro}

\IEEEPARstart{H}{andwritten} signature is a form of personal identification widely accepted, both socially and legally, and it has been used for centuries to authenticate documents such as bank checks, letters, contracts and many other that require proof of authorship. By signing, a person may provide unique information regarding the way she or he converts gesture intentions into spontaneous hand movement. Writing speed, traversed path, pen tilt, pressure applied, all these data are articulated to result in a static figure on signed documents \cite{nel2005}.

Signature analysis can be divided in two categories: offline and online. In the offline mode, either signatures are available through the traditional {\em wet ink} method (such as in paper documents), or they are available in scanned form, through optical devices, such as scanners and digital cameras. In both cases, all available data corresponds to static signature images. This type of approach is not the most efficient for verifying signatures due to the fact that relevant dynamic information is discarded.

In the online mode, a person uses a digitizing device (e.g. digitizing tablets or touchsceen devices) to directly record signals from the hand movement. This provides much more information than a static image, for the digitizing device typically can record several complementary signals, such as the path travelled by the pen tip, as well as its instantaneous speed, applied pressure and pen tilt. This approach is the one that dominates research on signature verification now, due to the worldwide spreading  of affordable acquisition devices \cite{plamondon}. 

However, the offline approach still has some attractive aspects. For instance, even today, many contracts and credit card authorization are performed through traditional signatures on paper. Indeed, although online signature verification has higher reliability, in many practical situations, for economical or practical reasons, {\em wet ink} signatures are yet useful biometric signals. And even in the unlikely scenario of a complete substitution of {\em wet ink} signatures by electronically acquired ones, at least the task of signature verification from ancient ink on paper documents should remain a relevant topic, due to the large amount of old signed documents, whose authenticity are potentially waiting to be verified \cite{qiao2006}.

To give some fundamental definitions and jargon, we assume that a signature verification is a process that determines whether a tested signature was produced by a target individual, from which at least one genuine signature is available. If under some chosen criteria the tested signature is similar to the genuine references, below a pre-established {\em similarity threshold},  it is labelled {\em true}, or a genuine signature. Otherwise,   the signature is labelled {\em false}, or a {\em forgery}. Moreover, Coetzer et al. \cite{coetzer} classify forgeries as
\begin{itemize}
\item \textit{Random forgery}: The forger does not know the author's name neither the original signature. Thus the false signature is completely random.
\item \textit{Simple forgery}: The forger knows the author's name, but she/he does not have access to the original signature.
\item \textit{Skilled forgery}: The forger has access to samples of genuine signatures, and also knows the name of the author. It can also be divided into two classes: \textit{Amateur} and \textit{Professional}. The \textit{Professional Skilled Forgery} is produced by a person with professional expertise in handwriting analysis, being able to produce a higher quality forgery than the \textit{Amateur}. 
\end{itemize}

In general, the offline signature verification process can be divided into four steps \cite{al2011}: \textit{Acquisition, Preprocessing}, \textit{Feature Extraction} and \textit{Comparison}. In the \textit{Preprocessing} step, image quality is improved and pixels are transformed to reduce the computational burden of the subsequent steps. Examples of techniques applied in this step are: thinning, color conversion, noise reduction, smoothing, morphological operations and resizing. For instance, Shah et al. \cite{shah2016} cropped images to exclude redundant white regions. 
The \textit{Feature Extraction} step is where most works propose innovations. According to Batista et al. \cite{batista2008}, an ideal feature extraction technique extracts a minimal feature set that maximizes interpersonal variability amongst signature samples from various subjects, whereas it minimizes intrapersonal variability amongst samples belonging to the same subject. Lee and Pan \cite{lee} divide the features into three classes: \textit{Global Features, Local Features} and \textit{Geometrical Features}.

Typical features extracted from offline signatures are marginal projections. Shanker and Rajagopalan \cite{shanker2007} extracts vertical projection of bitmaps corresponding to signatures, thus yielding profiles which are compared through Dynamic Time Warping (DTW). Likewise, Coetzer et al. \cite{coetzer} pushes a bit further the same idea, by using many marginal projections of the same signature, over different angles, what they call {\em Discrete Radon Transform}, whose behaviour is modelled with a Hidden Markov Model. Nguyen et al. \cite{nguyen2009} also use similar projections. Indeed, they use two techniques for global features extraction: the first is derived from the total {\em energy} a writer uses to create a signature, whereas the second technique employs information from the vertical and horizontal projections of a signature, focusing on the proportion of the distance between key strokes in the image, and the height/width of the signature. 

Although marginal projections are more commonly used in literature, straightforward approaches to feature extraction may also rely upon  image skeletonization \cite{gonzalez}. Typically, skeletonization is used to filter foreground pixels in bitmaps. But it can also be used to map offline signatures into sets of points, similar to online representations, which is appealing because online verification techniques may be deployed, such as the use of DTW to compare segments of points from different signatures. Indeed, this straightforward approach corresponds to the baseline method implemented in this paper, as explained in Section \ref{baseline}.

Once features are available, signature authenticity verification can be performed. To simulate actual verification, most academic works randomly select a small number of genuine signature samples from each user (typically from 5 to 15) to play the role of a set of enrolled signatures. Then, samples from the remaining dataset of false and genuine signatures are randomly taken to simulate verification attempts. These test samples are compared to the enrolled samples, and a decision is made. If a genuine signature is rejected, it is called a {\em false rejection} error. By contrast, if a forgery is accepted, it is called a {\em false acceptance} error. The experimental protocol for False Acceptance Rate (FAR) and False Rejection Rate (FRR) computation used in this work is explained in Section \ref{resultados}.   
 
Furthermore, in this paper, we propose a method inspired by online approaches, through the compact codification of segments of skeletonized offline signatures, as explained in Section \ref{proposto}. These skeletonized segments are the basic aspect of this work, and they are detailed in Section \ref{USCC}. Moreover, the  segmentation used in this work induces a straightforward method similar to classical online verification strategies, through DTW.   

As for our experiments, we take into account that, in biometrics, comparing performances measures using different databases can be misleading, which we consider an important issue. Therefore, to allow a direct comparison under the same database and the same experimental protocol for all compared methods, we use the well-known MCYT-100 online signature database, where each online sample was converted to an offline (bitmap) representation, as explained in Section \ref{resultados}, where comparative results are provided along with our experimental setup. It is worth noting that human baseline performance is presented for the same dataset, in the spirit of works such as \cite{coetzer2006} and \cite{Morocho2016}. Finally, in Section \ref{conclusoes}, we conclude by discussing our results and the usefulness of Mean Opinion Scores (MOS) as a potential goal to automatic verification performances.  
\section{Unambiguously connected skeleton segments}
\label{USCC}

Raw online signature signals are frequently represented by two vectors of samples: a sequence of regularly sampled horizontal positions, $x_{ONLINE}(n)$, and another sequence of corresponding vertical positions, $y_{ONLINE}(n)$, where $n$ stands for sample counter through time. As compared to offline representation, signature verification through signals $x_{ONLINE}(n)$ and $y_{ONLINE}(n)$ is significantly better. 

\begin{figure}[htb]
\centering{\includegraphics[width=90mm]{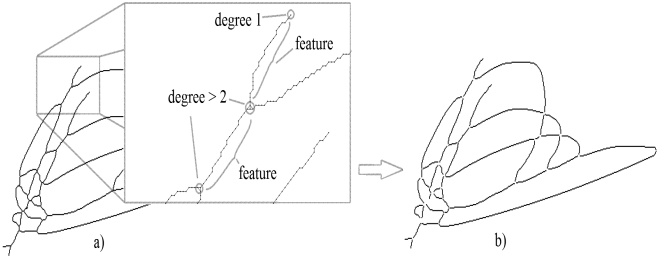}}
\caption{(a) Inside the circles one finds the points of the skeleton that delimits UCSS. (b) Full skeleton as a set of extracted features(UCSS).}
\label{fig2}
\end{figure}

Although we know that velocity information may not be completely recovered from offline representations, we address the offline signature verification problem by first recovering horizontal and vertical signals, which may be regarded as pseudo-versions of  $x_{ONLINE}(n)$ and $y_{ONLINE}(n)$. This is done through standard skeletonization, as described in \cite{gonzalez}. Unlike true online representations, skeletons from each signature are sets of unordered points. For instance, as in Figure \ref{fig2}, most offline signature skeletons can be regarded as sets of unordered pixels in a bitmap, even though some subsets of these pixels form segments that are clearly created in an unambiguous sequential hand gesture. 

Comparisons between online signatures are straightforward, because points $(x_{ONLINE},y_{ONLINE})$ are ordered through time. Analogously, the comparison between two offline signatures may also be done through the comparison of sequences of points $(x_{OFFLINE},y_{OFFLINE})$, representing black pixel coordinates in skeletons. However, ambiguities concerning the ordering of points turns this task into a combinatorial optimization problem whose computational cost may be prohibitive. 

To significantly reduce this cost, both methods proposed in this paper decompose offline signatures skeletons into {\em Unambiguously Connected Skeleton Segments} (UCSS), as illustrated in Figure \ref{fig2}. To define UCSS, an offline signature skeleton is regarded as connected graph $G(V,E)$ where the vertices $V$ are the points of the skeleton and the edges $E$ are  bidirectional connection between neighbouring points (8-connected neighbourhood). We also consider that the degree of a vertex $V$ is the number of neighbouring vertices that $V$ is connected to. Therefore, each UCSS is a sequences of directly connected vertices found between:
\begin{itemize}
\item[a)] two vertices with degrees greater than 2 (internal segments),
\item[b)] an one-degree vertex and a vertex with degree greater than 2 (extremities), or
\item[c)] two one-degree vertices (isolated lines). 
\end{itemize}

We assume that each segment, $s_{m,n,k}$, is a portion of signature where points are unambiguously ordered, apart from a single ambiguity in the overall direction of the pen movement (i.e. one does not know in what end of the UCSS the movement of the pen begins). Thus, the $n$-th signature sample, $n=1,2,\ldots,N$, from the $m$-th signer, $m=1,2,\ldots,M$ is represented by a set $\{s_{m,n}\}=\{s_{m,n,1},s_{m,n,2},\ldots, s_{m,n,K}\}$, where $s_{m,n,k}$ is the $k$-th UCSS, or a sequence of $L_{m,n,k}$ pairs of coordinate points, $(x_i,y_i)$, $1 < i \leq L_{m,n,k}$.  
 
Moreover, to take into account the single ambiguity in the overall direction of the pen movement, each UCSS is represented twice: first with the sequence of pairs in a given order, $s_{m,n,k}=[(x_1,y_1), (x_2,y_2), \ldots, (x_{L_{m,n,k}},y_{L_{m,n,k}})]$, and then in reversed order, $s_{m,n,k}^*=[(x_{L_{m,n,k}},y_{L_{m,n,k}}), (x_{L_{m,n,k}-1},y_{L_{m,n,k}-1}),\ldots,(x_1,y_1)]$.

\section{Baseline Method} \label{baseline}
Two methods for automatic offline signature verification are proposed in this paper. The first method is considered as a baseline, for it is a straightforward application of Dynamic Time Warping to compute distances between UCSS. In this method, the standard DTW method under Itakura's restrictions \cite{itakura} is applied to systematically compare every segment $s_{m,n,k}$ (and its reverse, $s_{m,n,k}^*$) to every segments of a given {\em bag of segments}, extracted from reference signatures. Consider, for instance, a test signature, $\{ s_{m,test} \}$  with $K_{test}$ UCSS, and a bag of segments, $\{ B \}$, with $K_{B}$ UCSS (i.e. all segments, from all references, are merged into the single set $\{ B \}$). Then, each UCSS in the test signature is compared to all $K_{B}$ UCSS from $\{ B \}$, and the minimum distance is taken. In other words, each UCSS in a test signature is associated to the single UCSS in $\{ B \}$ which yields the minimum DTW distance. More precisely,
\begin{equation}\label{custoComDTW}
C_{m,test}(k,\{B\}) =  \frac {1}{L_{m,test,k}} \min\limits_{j}DTW(s_{m,test,k},s_{B,j}),
\end{equation}
\noindent and the average distance between sets $\{ s_{m,test} \} $ and $\{B\}$ is given by
\[
{\bar C}_{m,test}(\{B\}) = \frac{1}{K_{test}} \sum\limits_{k=1}^{K_{test}} C_{m,test}(k;\{B\})
\]
\noindent where $DTW(s_{m,test,k},s_{B,j})$ stands for Dynamic Time Warping distance under Itakura's restriction between $s_{m,test,k}$ and $s_{B,j}$, or its reversed version, $s_{B,j}^*$, depending on which one yields the lowest distance.  Moreover, $k \in{1,2,\ldots,K_{test}}$, $j \in{1,2,\ldots,K_B}$, and $test$ is a pointer to a signature from the test set. Moreover,  $L_{m,test,k}$ is the length (number of points) of the $k$-th UCSS, of the tested signature. 

In this work, we randomly take 5 genuine signatures from each individual as reference set, denoted as $\{s_{m,ref_1}\}$,$\{s_{m,ref_2}\}$,..., $\{s_{m,ref_5}\}$. Therefore, $\{ B \}$ is the union of all 5 references, namely: $\{ B \}=\{\{s_{m,ref_1}\} \bigcup \{s_{m,ref_2}\} \bigcup \ldots \bigcup \{s_{m,ref_5}\}\}$, and $K_{B}$ is the resulting cardinality of $\{ B \}$. 

To provide a better score for tested signatures, we also define 5 partial bags of segments, $\{ B \}_{ref_i}$, where segments of the $i$-th reference signature are excluded from $\{ B \}$. As a result, we are able to compute an average distance between each reference signature, $\{s_{m,ref_i}\}$, and the corresponding remaining bag of segments, $\{ B \}_{ref_i}$, as follows
\[
{\bar C}_{0}(\{B\})=(1/5) \sum\limits_{i=1}^{5} {\bar C}_{m,ref_i}(\{ B \}_{ref_i}).
\]     

Finally, the total distance between a tested signature $\{s_{m,test}\}$ and a genuine set of references, summarised by the bag of segments $\{ B \}$, is defined as:
\begin{equation}\label{eqJ}
J_{m,ref}=\frac{{\bar C}_{m,test}(\{B\})}{{\bar C}_{0}(\{B\})},
\end{equation}    
\noindent where ${\bar C}_{0}(\{B\})$ plays the role of a normalization score. 
 
An important drawback of this baseline method is the computation of more than $K_{B} \times K_{test}$ DTW distances in order to obtain a single score/cost for each tested signature, where $K_{B}$ is roughly 5 times the average number of segments (UCSS) in a genuine signature. As a consequence, this method has a high computational burden.     

\section{Proposed method with UCSS subsampling}\label{proposto}
\begin{figure}[!h]
\centering{\includegraphics[width=85mm]{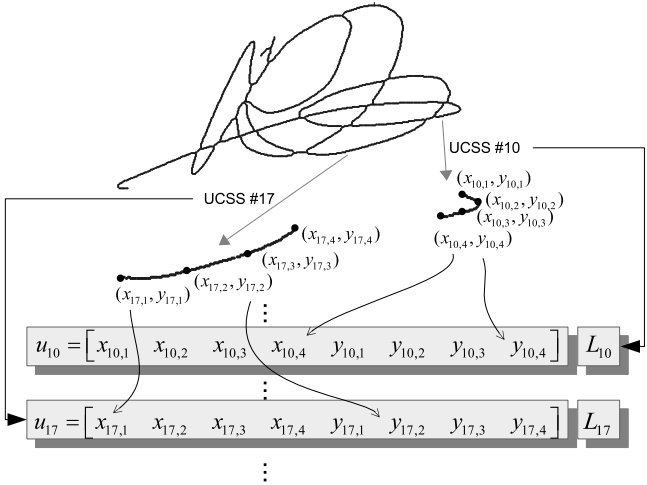}}
\caption{Proposed method: illustration of 8D vectors and lengths of two segments (segments 10 and 17). Each segment (UCSS) is coded with $8+1=9$ scalars. The last scalar represents the segment length.}
\label{fig3}
\end{figure}

To significantly alleviate the computational load of the baseline method, we encode each UCSS as an eight-dimensional (8D) vector that roughly represents the shape and the position of the encode UCSS, plus a scalar corresponding to the UCSS length (the length is given in terms of number of points in the UCSS), as illustrated in Figure \ref{fig3}. 

This UCSS encoding strategy is the main aspect of the proposed method. We assume that almost all UCSS are short enough to prevent strong warping, therefore, one may get rid of the high DTW computation cost by replacing each UCSS with 4 subsampled points. In other words, it is assumed that UCSS comparisons through DTW are almost equivalent to the much faster Euclidean distance computation between the corresponding 8D vectors (UCSS length is not taken into account when plain DTW is used). Indeed, a UCSS $s_{m,n,k}$ can be regarded as a composition of two sampled signals, say $x_n$ and $y_n$, with $n = 1,2, \ldots, L_{UCSS}$, and the proposed coding scheme just takes 4 equally spaced subsamples of $x_n$ and $y_n$, thus yielding an 8-dimensional vector $u_{m,n,k}$. For practical purposes, we then assume that:
\[
\frac{DTW(s_{m,n,i},s_{m,n,j})}{L_{m,n,i}} \approx \frac{D_2(u_{m,n,i},u_{m,n,j})}{4} ,
\] 
\noindent where $D_2(u_{m,n,i},u_{m,n,j})$ is the accumulated squared Euclidean distance between the 4 corresponding points in the two compared UCSS. We highlight that, as for the definition of function $DTW$, in Equation \ref{custoComDTW}, $D2$ is also defined as the minimum of two distances, where both $u_{m,n,j}$ and its reversed version, $u_{m,n,j}^*$ are considered. For the non-reversed $u_{m,n,j}$ only, we obtain $D_2(u_{m,n,i},u_{m,n,j}) =\sum\limits_{k=1}^8 ( u_{m,n,i}(k) - u_{m,n,j}(k) )^2$. Therefore, the comparison between two signatures is significantly simplified through the use of the following distance, as compared to that in Equation \ref{custoComDTW}:
\begin{equation}\label{custoSemDTW}
C_{m,test}(k,\{U\}) =  \frac {1}{4} \min\limits_{j}D_2(u_{m,test,k},u_{U,j}),
\end{equation}
\noindent where $\{U\}$ is the set of all 4-points segments from the reference signatures. Analogously, every subset $\{ B \}_{ref_i}$ is replaced with $\{ U \}_{ref_i}$, and the final score/cost for a given signature can be computed as in Equation \ref{eqJ}.
 
\section{Experimental Results}\label{resultados}

In this work, we use the well-known MCYT-100 online signatures database, for which error rates (for online verification task) can be abundantly found in literature \cite{ferrer2012,shah2016,hafemann2015}. This database is a subcorpus of the MCYT database \cite{mcyt}, acquired from $M=100$ different writers. Each writer provided 25 genuine signatures, whereas 5 different volunteers provided 25 skilled forgeries per signature ($N=50$). All signatures were acquired through a WACOM Intuos A6 USB Tablet at constant sampling rate of 100 Hz. 

To obtain the offline signatures we need for our experiments, we first convert each MCYT online sample into an image where only horizontal ($x_{ONLINE}$) and vertical ($y_{ONLINE}$) pen tip positions through time are considered, as follows: 
\begin{itemize}
\item[(i)] Points in each online signature are interpolated using splines to allow for oversampling of the otherwise sparse representation (due to the relatively low sampling rate of 100 samples per second). 
\item[(ii)] The oversampled set of points are numerically rounded to integer values, and 
\item[(iii)] they are also dilated until segments are approximately four pixels wide. 
\end{itemize}
These three steps are enough to convert the entire online data into offline signatures. Indeed, the resulting images are used in the Mean Opinion Score experiment detailed in Subsection \ref{MOS}. As for the other experiments, another sequence of steps are still taken to convert offline signatures into pseudo-online ones, namely:

\begin{itemize}
\item[(i)] The standard skeletonization method described in \cite{gonzalez} is applied to each dilated signature image. 
\item[(ii)] Resulting skeletons, or sets of 2D points from each signature sample are centered at the origin, whereas their variance in both vertical and horizontal directions are scaled to one.
\end{itemize}
\begin{figure}[htb]
\centering{\includegraphics[width=80mm]{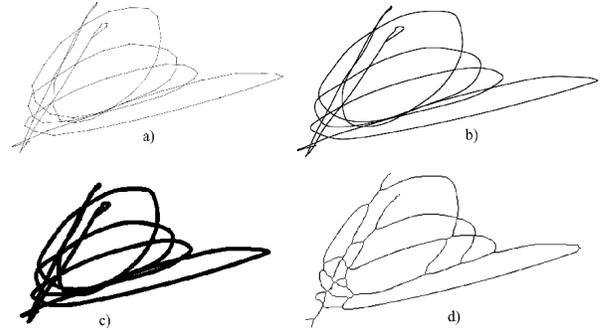}}
\caption{(a) Online signature. (b) Oversampled online signature. (c) Dilated image -- approximately four pixels wide lines. (d) Skeleton of the signature.}
\label{fig1}
\end{figure}

From the resulting pseudo-online versions of the signatures, represented by a set of coordinate pairs, UCSS are extracted. It is noteworthy that the order in which these pairs are presented no longer stands for a discrete time counter, as in true online representation, for it now represents a mere skeleton point counter, whose correspondence to time ordering is unknown.   

In all experiments, to simulate an actual biometric system, we randomly choose 5 genuine signatures from the database to form the enrolment card. Afterwards, the other signatures of the same user are randomly sampled and compared with those into the enrolment card. Given a decision threshold, the proportion of true signatures whose costs are above this threshold (thus wrongly rejected as not genuine) is the estimated FRR, whereas the proportion of false signatures whose distances are below the threshold is the estimated FAR.

\begin{figure}[htb]
\centering{\includegraphics[width=85mm]{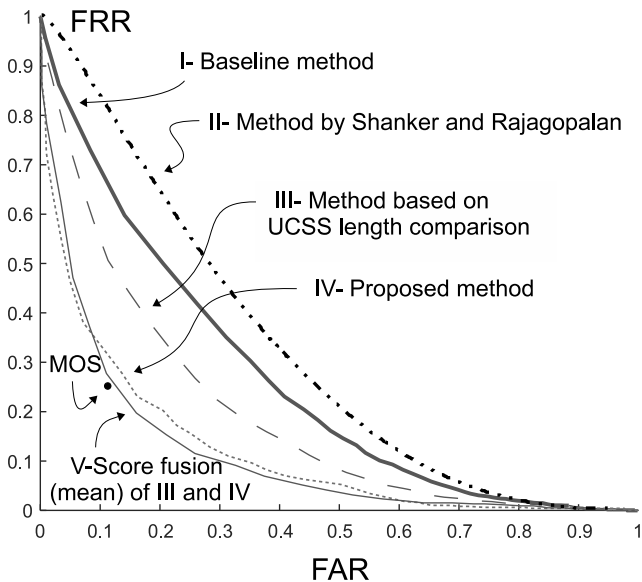}}
\caption{ROC curves for all tested methods. In this set of experiments, we use 5 randomly chosen reference signatures to simulate enrolment, and a pool of 20 remaining genuine signatures, along with 25 false ones to test the simulated system. A black dot also indicates the MOS performance, as explained in Subsection \ref{MOS}, for a subjective decision threshold that cannot be handled to yield a ROC.}
\label{figROC}
\end{figure}

We also compared the computational burden of the two proposed methods to process and compare one enrolment card, with 5 genuine samples per signature, and to compute scores for all $20+25=45$ remaining signature samples per reference. The processing time for the baseline method was thousands times greater than that for the proposed method with UCSS subsampling. Surprisingly, the lighter method yielded a significantly better performance, as presented in Figure \ref{figROC}. 

An interesting by-product of both methods is that after two UCSS are associated (according either to Equation \ref{custoComDTW} or to Equation \ref{custoSemDTW}), instead of comparing their shapes, one may just compare their lengths, thus yielding a new score corresponding to the absolute difference of associated lengths. This can be regarded as a third method, here referred to as the {\em length based} one, in Figure \ref{figROC} and Table 1.  

To provide a wider comparison scenario for the proposed methods based on UCSS, we also included in the comparison experiments the method published by \cite{shanker2007}, which extracts projections of bitmaps corresponding to signatures, and compare them through a modified DTW, where so called {\em stability measures} are included to improve performances. We reproduced the best implementation of their method (as explained in their paper), and we applied it to the the same offline signatures we used to test our methods, under the same experimental protocol. We highlight that although the method by \cite{shanker2007} was implemented and revised to strictly follow their instructions, we only apply it to the database of pseudo-offline signatures, as illustrated in Figure \ref{fig1}. Moreover, instead of 10 reference genuine signatures, as in \cite{shanker2007}, we use 5 ones to assure the same protocol to all compared methods.  

In a second round of experiments, we repeated 50 times, for each method, the following protocol:  5 genuine signatures were randomly chosen as references (enrolment), whereas a set of 10 (5 from the remaining genuine signatures, plus 5 randomly chosen false ones) were used to test the simulated system. For each independent trial, we adjusted the threshold decision to obtain the operational point where FAR equals FRR, the Equal Error Rate (EER). Table 1 presents all average results per method, in terms of EER and its standard deviation over the 50 independent trials.
\begin{table}\label{TabResults}
\centering{
\caption{Results for each method after 50 independent runs. Each run corresponds to a random partition of 5 reference and 10 test signatures (5 genuine ones). }
\begin{tabular}{|c|c|c|}
\hline 
Method & EER & std. dev. \\ 
\hline 
I-Baseline & 32.2 \%     & 1.4 \% \\ 
\hline 
II- Method by \cite{shanker2007} &  36.7  \% & 0.6 \% \\ 
\hline 
III-Length based & 25.1 \% & 1.2 \% \\ 
\hline 
IV- Shape based (proposed) & 19.4 \% & 1.3 \% \\ 
\hline 
V-Fusion (mean) of scores &  &  \\ 
from III and IV & 18.7 \% & 1.0 \% \\ 
\hline 
\end{tabular}} 
\end{table}

In both sets of experiments, scores of the methods III and IV are also fused through simple arithmetic mean, yielding an improved performance, as shown in Figure \ref{figROC} and Table 1.

\subsection{Mean Opinion Score Extraction Protocol}\label{MOS}

\begin{figure}[htb]
\centering{\includegraphics[width=85mm]{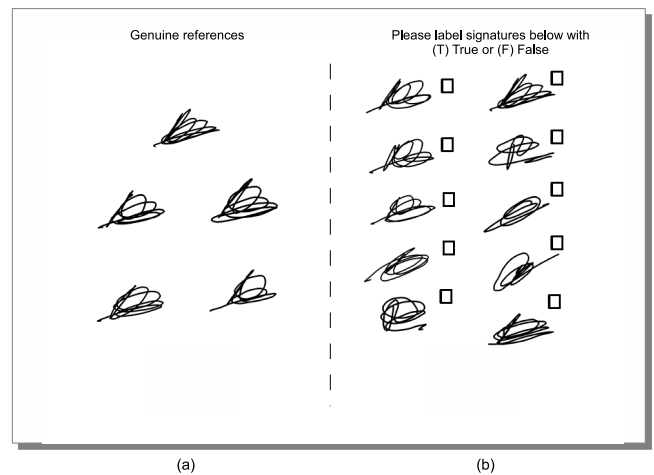}}
\caption{(a) Reference panel with five randomly chosen genuine signatures (b) Panel with five randomly chosen forgeries and five randomly chosen genuine signatures.}
\label{figCards}
\end{figure}

To quantify the human performance for the same task, we also prepared 239 two-page cards, each one corresponding to a genuine source of signatures -- a genuine signer -- in the MCYT database. These cards contain five genuine signatures on its left side, and ten signatures randomly chosen in its right side. Only 5 out of the 10 signatures on the right side are genuines. An example of these cards can be seen in Figure \ref{figCards}. 
These cards were presented to different students and lecturers in our university (willing volunteers), and these volunteers were carefully instructed to study the genuine signatures presented on the left part of the page, and then to label all signatures on the right part, by writing in the boxes next to each signature a T, for true (genuine), or a F, for false. No further {\em a priori} was provided. We highlight that the volunteers did not know the proportions of true and false signatures in each card. 

A total of 239 cards were filled by 103 volunteers, and after comparing all provided labels to the true {\em hidden} labels, the estimated Mean Opinion Score (MOS) obtained was consolidated as:
\begin{itemize}
\item  25.6\% of False Rejection Rate  and 
\item  11.0\% of False Acceptance Rate.
\end{itemize}
The dot in Figure \ref{figROC} allows the visual comparison between the rates FAR and FRR from MOS (to which the decision threshold cannot be known or handled) and the ROC curves from the automatic methods, for a range of possible decision thresholds.  

\section{Conclusion}\label{conclusoes}

This brief work was inspired by the evident superiority of  online signature verification methods, as compared to offline ones. Therefore all methods proposed here are based on skeletonization, possibly the most straightforward method to obtain pseudo-online signature representations from images. The baseline method cross-compares all unambiguously connected sequences of points from skeletons, where the conception of a Unambiguously Connected Skeleton Segment (UCSS - whose formal definition was given in Section \ref{USCC}) plays a pivotal role. By assuming that UCSS shapes and position are relevant information for biometric verification, we should expect that the systematic cross-comparison of UCSS, through DTW, would yield the best performance, at a high computational cost.

However, it was noticed instead that the alternative method initially proposed to alleviate the high computational burden of the baseline method -- by considering only 4 points per UCSS -- yielded a significantly better performance, as compared to the baseline method. Moreover, even the very simple by-product method based on the comparison between UCSS lengths performed better than the baseline method. Indeed, the baseline method only  outperformed the method by \cite{shanker2007}, which uses an improved DTW, but relies upon marginal projections of signatures instead of segments like the UCSS proposed in this work. 

From these results, we conjecture that, in average, UCSS is a good segmentation option, but UCSS shape details are not relevant information carriers for the biometric verification purpose. Indeed, given the comparative performances, we should even conclude that UCSS shape details are disturbing noises, for biometric verification task. This point is an interesting matter for future works. 

Regarding the by-product method based on UCSS length, it is noteworthy that the use of either DTW or Euclidean distance to match UCSS is a necessary step. In other words, behind the apparent simplicity of this method, one should be aware that the matching of UCSS is a not so simple step of it, which also rises interesting questions. For instance, the superiority of the joint approach -- where UCSS shape and length are combined -- may have a connection to the {\em lost} signal of pen tip velocity, which in turn is the main signal for biometric verification in \cite{canuto2016}. Indeed, UCSS shape (straightness) and length are expected to be somehow dependent on pen tip velocity, either through two-thirds power law, or through isochrony \cite{viviani1995}, and this dependency is also an attractive subject for further works. For now, in this letter, we just conjecture that the fusion of length and shape based scores is somehow related to inferred velocity signal, given a signature image, which may explain its relatively good performance. 

Clearly, even the best performance presented here is far from performances typically found in literature for online verification, with the same database in its original (online) form. By contrast, in this work we assume that, for the offline verification task, the best possible performances are not far from what a crowd of willing and attentive humans can do. Therefore, we choose to use a MOS-based rate as basis for comparison, and it indicates that the best performances we obtained are indeed close to that of a crowd of humans. Nonetheless, further experiments using actual offline signature databases and other competing methods are intended to be done in the sequel of this work.

% use section* for acknowledgment
\section*{Acknowledgment}
This work was supported by grant from the Conselho Nacional de Desenvolvimento
Cient\'ifico e Tecnol\' ogico (CNPq) to J.M.

\ifCLASSOPTIONcaptionsoff
  \newpage
\fi

\end{document}